\documentclass[runningheads]{llncs}
\usepackage{graphicx}
\usepackage{amssymb}
\usepackage{amsmath}
\usepackage{subfigure}
\usepackage{caption}
\begin{document}
\title{Relation Modeling with Graph Convolutional Networks for Facial Action Unit Detection}
\titlerunning{Relation Modeling with GCNs for Facial Action Unit Detection}
%

\author{Zhilei Liu\inst{1} \and
Jiahui Dong\inst{1} \and
Cuicui Zhang\inst{2}\thanks{Corresponding author.} \and
Longbiao Wang\inst{1}\and
Jianwu Dang\inst{1,3}}

\authorrunning{Z. Liu et al.}
%
\institute{College of Intelligence and Computing, Tianjin University, China\\
\email{\{zhileiliu, 2118216008, longbiao\_wang\}@tju.edu.cn} \and
 School of Marine Science and Technology, Tianjin University, China
\email{cuicui.zhang@tju.edu.cn} \and
School of Information Science, JAIST, Japan\\ \email{jdang@jaist.ac.jp}
}
%
%
\maketitle              
\begin{abstract}
\sloppy
 Most existing AU detection works considering AU relationships are relying on probabilistic graphical models with manually extracted features. This paper proposes an end-to-end deep learning framework for facial AU detection with graph convolutional network (GCN) for AU relation modeling, which has not been explored before. In particular, AU related regions are extracted firstly, latent representations full of AU information are learned through an auto-encoder. Moreover, each latent representation vector is feed into GCN as a node, the connection mode of GCN is determined based on the relationships of AUs. Finally, the assembled features updated through GCN are concatenated for AU detection. Extensive experiments on BP4D and DISFA benchmarks demonstrate that our framework significantly outperforms the state-of-the-art methods for facial AU detection. The proposed framework is also validated through a series of ablation studies.

\end{abstract}
\vspace{-2mm}
\section{Introduction}
\sloppy
In recent years, research on Facial Action Unit (AU), as a comprehensive description of facial movements, has attracted more and more attention in the field of human-computer interaction and affective computing. Facial AU detection is beneficial to facial expression recognition and analysis. According to statistical calculation and facial anatomy information, strong relationships are exist among different AUs under different facial expressions, e.g., happiness might be the combination of AU12 (Lip Corner Puller) and AU13 (Cheek Puffer). 

Most of existing AU detection methods focus on AU relationship modeling implicitly. For example, probabilistic graphic models including Bayesian Networks~\cite{8510873}, Dynamic Bayesian Networks~\cite{Zou2005A} and Restrict Boltzmann Machine~\cite{Wang2014Capturing} have demonstrated their effectiveness of relation modeling for AU detection. However, these generative models are always integrated with manually extracted feature, i.e. LBP, SIFT, HoG, which limits its extension ability with state-of-the-art deep discriminative models.

\indent
With the recent development of deep graph networks, relation modeling with graph based deep graph models has attracted more and more attention. In this paper, we use the graph convolutional network (GCN)~\cite{kipf2016semi} for AU relation modeling to strengthen the facial AU detection. In particular, reference to EAC-Net~\cite{li2017eac}, AU related regions are extracted at first, these AU regions are feed into some specific AU auto-encoder for deep representation extraction in the next. Moreover, each latent representation is pull into GCN as a node, the connection mode of GCN is determined by the relationship of AUs. Finally, the assembled features are concatenated for AU detection. These auto-encoders are trained firstly, then the whole framework is trained together.\\
\indent
The contributions of this paper are twofold. (1) We propose a deep learning framework for AU detection with graph convolutional network for AU relation modeling. (2) Results of extensive experiments conducted on two benchmark datasets demonstrate that our proposed framework significantly outperforms the state-of-the-art.

\vspace{-2mm}
\section{Related Work}
\sloppy
Our proposed framework is closely related to facial AU detection and graph convolutional network.\\

\textbf{Facial AU detection}: Based on previous research\cite{Zou2005A,Wang2014Capturing,song2015exploiting}, AUs are in contact, which make itself a problem different from standard expression recognition. To capture such correlations, a generative dynamic Bayesian networks (DBN)~\cite{Zou2005A} was proposed to model the AU relationships and their temporal evolution. Rather than learning, pairwise AU relations can be explicitly inferred using statistics in annotations, and then injected such relations into a multi-task learning framework to select important patches for each AU. In addition, a restricted Boltzmann machine (RBM)~\cite{Wang2014Capturing} was developed to directly capture the dependencies between image features and AU relationships. Following this direction, image features and AU outputs were fused in a continuous latent space using a conditional latent variable model. Song et al.~\cite{song2015exploiting} studied the sparsity and co-occurrence of AUs. Although improvements can be observed considering the relationships among AUs, these approaches rely on manually extracted features such as SIFT, LBP, or Gabor, rather than deep features.

With the recent rise of deep learning, CNN have been widely adopted to extract AU features. Zhao et al.~\cite{zhao2016deep} proposed a deep region and multi-label learning (DRML) network to divide the face images into 8 $\times$ 8 blocks and used individual convolutional kernels to convolve each block. Although this approach treats each face as a group of individual parts, it divides blocks uniformly and does not consider the FACS knowledge, thereby leading to poor performance. Wei Li et al.~\cite{li2017eac} proposed Enhancing and Cropping Net (EAC-Net), which intends to give significant attention to individual AU centers; however, this approach does not consider AU relationship modeling, and the lack of RoI-level supervised information can only give coarse guidance. All these researches demonstrate the effectiveness of deep learning on feature extraction for AU detection task. However, they all do not consider the AU relation modeling.\\
\indent
\textbf{Graph Convolutional Network}: There have been a lot of works for graph convolution, whose principle of constructing GCNs mainly follows two streams: spatial perspective~\cite{duvenaud2015convolutional,atwood2016diffusion,niepert2016learning} and spectral perspective~\cite{ng2018bayesian,bruna2013spectral,henaff2015deep,hammond2011wavelets,defferrard2016convolutional}. Spatial perspective methods directly perform the convolution filters on the graph vertices and their neighbors. Atwood et al.~\cite{atwood2016diffusion} proposed the diffusion-convolutional neural networks (DCNNs). Transition matrices are used to define the neighborhood for nodes in DCNN. Niepert et al.~\cite{niepert2016learning} extracts and normalizes a neighborhood of exactly k nodes for each node. And then the normalized neighborhood serves as the receptive field for the convolutional operation. Different with the spatial perspective methods, spectral perspective methods utilize the eigenvalues and eigenvectors of graph Laplace matrices. Bruna et al.~\cite{bruna2013spectral} proposed the spectral network. The convolution operation is defined in the Fourier domain by computing the eigendecomposition of the graph Laplacian.
\\
\indent
Recently, Li et al.~\cite{li2019semantic} proposed the AU semantic relationship embedded representation learning (SRERL) framework to combine facial AU detection and Gated Graph Neural Network (GGNN)~\cite{li2015gated} and achieved good results. But the commonly used Graph Convolutional Network (GCN) for classification task with relation modeling is adopted for AU relation modeling in our proposed method, while the Gated Graph Neural Network (GGNN) adopted in [13] is inspired by GRU and mainly used for the task of Visual Question Answering and Semantic Segmentation. In addition, our method has only about 2.3 million parameters, but SRERL[13] has more than 138 million parameters.
\\
\indent
In this paper, we apply the spectral perspective of GCNs~\cite{kipf2016semi} for AU relation modeling. Our GCN is bulit by stacking multiple layers of graph convolutions with AU relation graph. The outputs of the GCNs are updated features for each AU region node by modeling their relationships, which can be used to perform classification.

\vspace{-2mm}
\section{Proposed Method}

\begin{figure*}[htb]
\centering
\includegraphics[scale=0.35]{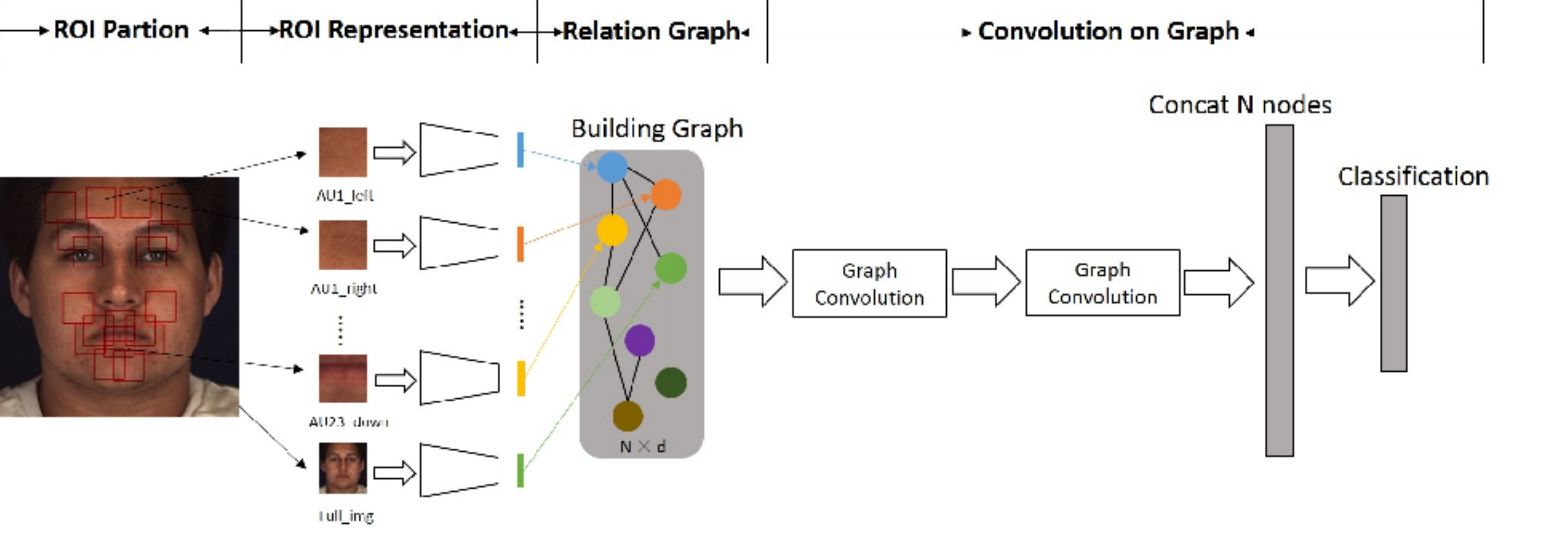}
\caption{The proposed AU-GCN framework for AU detection.}
\label{fig1}
\end{figure*}

\vspace{-2mm}
\subsection{Overview}\label{3.1}
The propose AU-GCN framework for AU detection by considering AU relation modeling through GCN is shown in Figure~\ref{fig1}, in which, four modules are included: AU local region division, AU local region representation, AU relation graph, Convolutions on graph. Given the face image with facial landmark key points, AU related local regions are extracted at first by taking EAC-Net~\cite{li2017eac} as a reference. After that, deep representations of each AU region are represented by the latent vectors of an auto-encoder supervised through the reconstruction loss and the AU classification loss. In the next, each latent vector is pull into GCN as a node, and AU relationships are modeled through the edges of GCN. Multiple layers of graph convolution operations will be applied on the input data and generating higher-level feature maps on the graph. It will then be classified by the modified multi-label cross entropy loss and Dice loss to the AU correct classification. We will now go over the components in the AU-GCN model as following. 

\vspace{-2mm}
\subsection{AU ROI Partition Rule}\label{AA}
The most recent deep learning based image classification methods make use of CNN for feature extraction, and the basic assumption for a standard CNN is the shared convolutional kernels for an entire image. For an image with the relatively fixed structure, such as a human face, a standard CNN may fail to capture those subtle appearance changes. In order to focus more on AU specific regions, the AU local region partition rules are defined at first by taking FACS~\cite{ekman1997face} and EAC-Net~\cite{li2017eac} as reference. 
\\
\indent
The first step is to use the facial landmark information to get the AUs centers. The landmark points provide rich information about the face, which help us to locate specific AU related facial areas. Then, taking this AU related landmark as the center to extract the n$\times$n size region as AU local region. Figure~\ref{fig21} shows the AU region partition of the face, in which, the face image is partitioned into 19 basic ROIs using AU related landmarks, AU12, AU14 and AU15 share a ROI, AU23, AU24, AU25 and AU26 share a ROI. These 12 ROIs are shared by different benchmark datasets, i.e. BP4D~\cite{zhang2013high} and DISFA~\cite{mavadati2013disfa} datasets, in which, 6 AU ROIs for BP4D dataset, and addition one for DISFA dataset. Due to the fact that previous ROIs are all the facial local feature, the facial global feature is ignored, another special ROI representing the whole face image in introduced. All these AU related ROIs will be resized into n$\times$n for further representation learning and relation modeling. Finally, BP4D dataset has 19 ROIs, DISFA dataset has 14 ROIs.

\vspace{-3mm}
\subsection{AU Deep Representation Extraction}
\begin{figure}[htbp]
\centering
\subfigure[AU ROI Partition.]{
\begin{minipage}[t]{0.5\linewidth}
\centering
\includegraphics[scale=0.35]{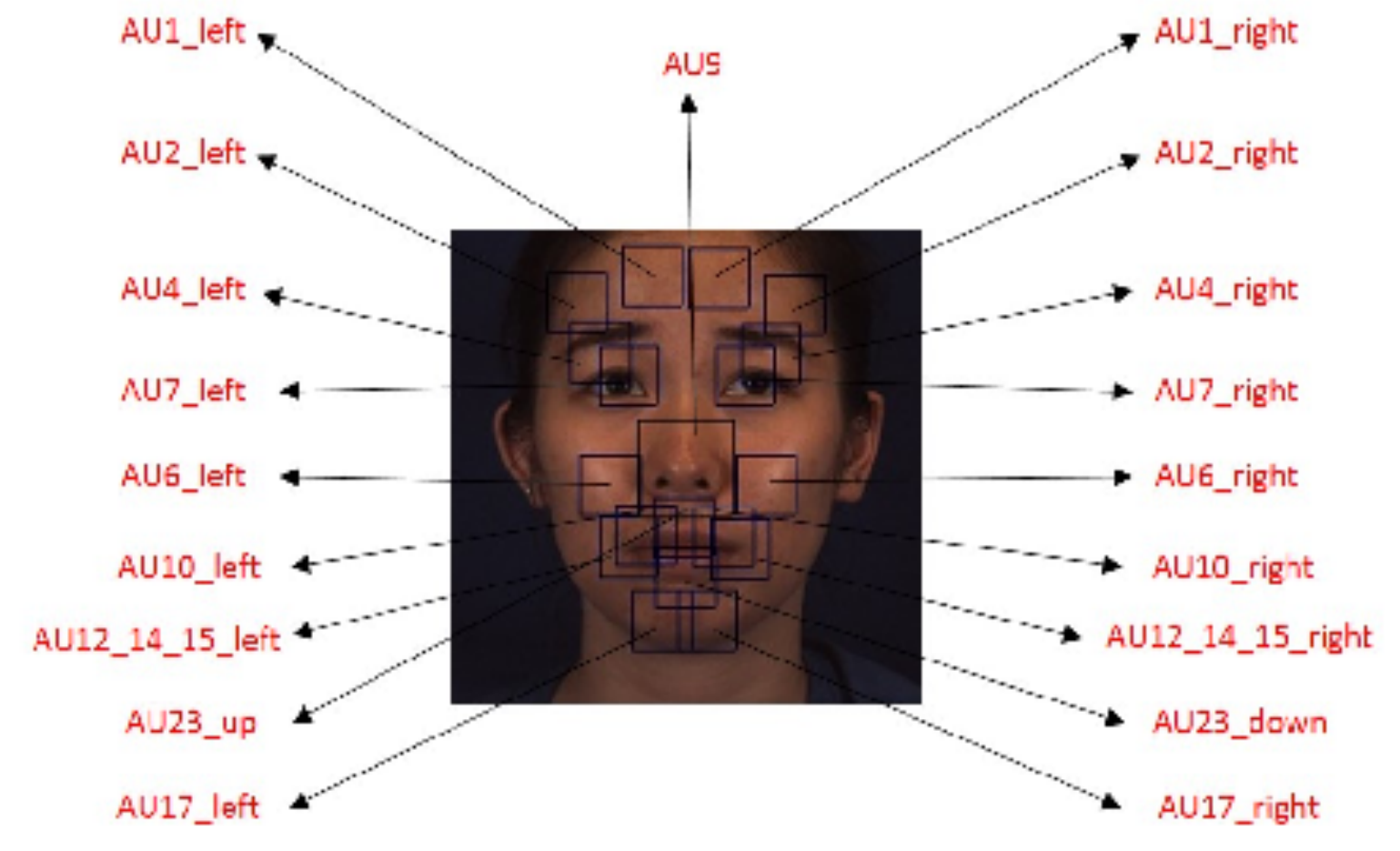}
\label{fig21}
\end{minipage}%
}%
\subfigure[AU representation extraction network.]{
\begin{minipage}[t]{0.5\linewidth}
\centering
\includegraphics[scale=0.35]{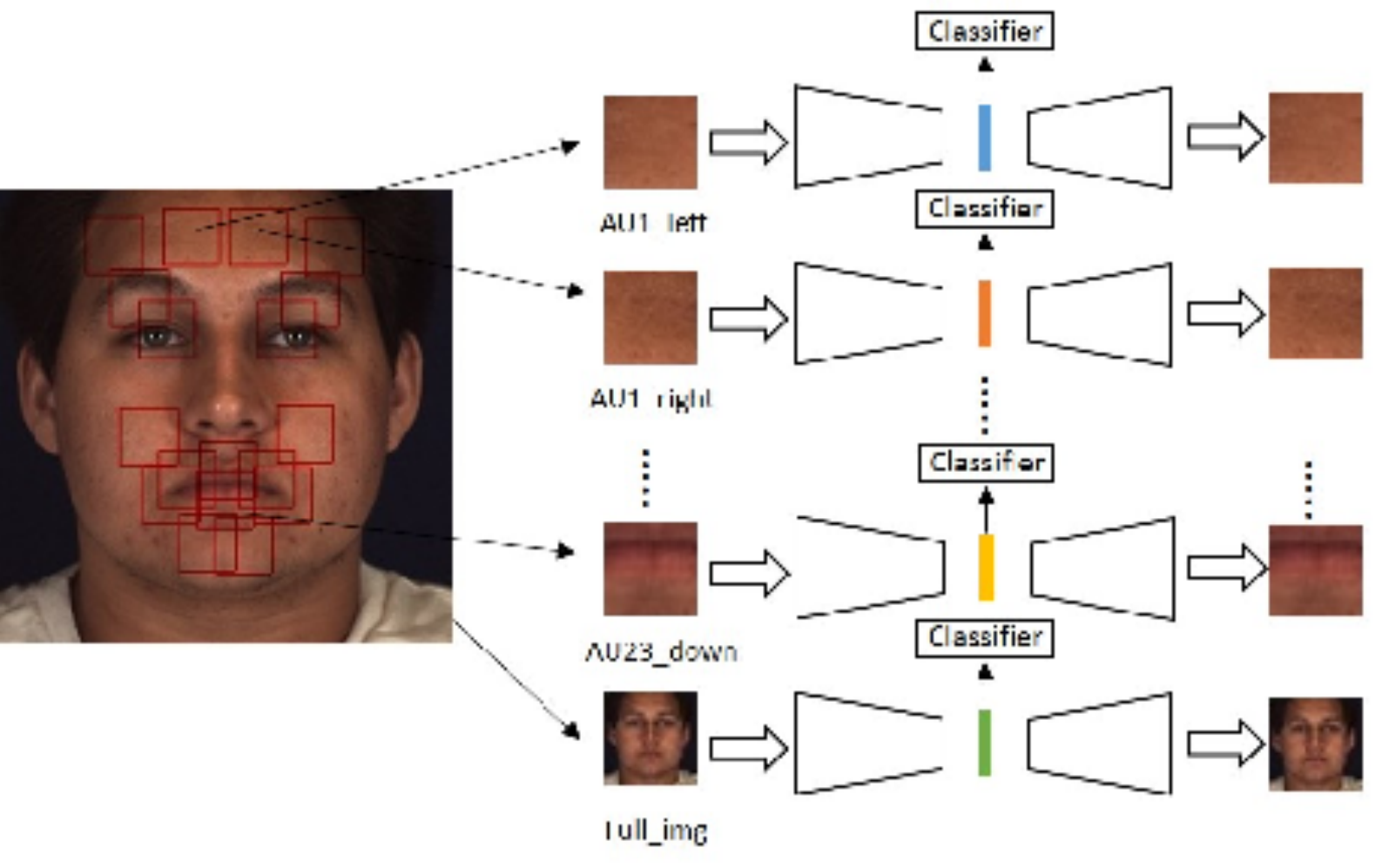}
\label{fig22}
\end{minipage}%
}%
\centering
\caption{AU ROI partition and AU deep representation extraction network.}
\end{figure}

\vspace{-3mm}
Figure~\ref{fig22} shows the architecture of network for AU deep representation extraction. This purpose of this step is to get $d_0$-dim deep representations full of AU information for further AU relation modeling and AU detection. 

\vspace{-3mm}
\vspace{-3mm}

\indent
The AU specific ROIs obtained in the previous step are feed into AU specific auto-encoders (AEs)~\cite{masci2011stacked} to reconstruct each AU ROI. To get latent vectors full of AU information, two kinds of losses are introduced here to constrain the extracted deep representations. The first loss is the pixel-wise L1-reconstruction loss $L_{R}$:
\begin{equation}
        L_{R}(I^{GT}, I^{R}))=\frac{1}{n^{2}}\sum_{i=1}^{n}\sum_{j=1}^{n}\left |I^{GT}_{ij}  - I^{R}_{ij}\right |
\end{equation}
\vspace{-2mm}
\\
\indent
where $n$ is the size of each AU ROI, $I^{GT}$ denotes the ground truth AU ROI image, $I^{R}$ denotes the reconstructed AU ROI image.

To make sure that the extracted AU deep representation contains as more AU information as possible, the second loss $L_{ROI\_softmax}$ for ROI-level multi-label AU detection is introduced as following:

\vspace{-3mm}
\begin{equation}
\begin{aligned}
L_{ROI\_softmax}&(Y^{ROI}_{r},\hat{Y}^{ROI}_{r}) =
\\&-\frac{1}{C}\sum_{c=1}^{C}\left [ Y_{rc}^{ROI}log\hat{Y}_{rc}^{ROI} + (1 - Y_{rc}^{ROI})log(1 - \hat{Y}_{rc}^{ROI}) \right]
\end{aligned}
\end{equation}
\vspace{-3mm}
\\
\indent
where $C$ is the number of the classes, $R$ is the number of the ROIs obtained in the previous step, i.e., 19 ROIs and 14 ROIs are defined in BP4D dataset and DISFA dataset respectively according to the provided AU labels, the ground truth of AU label is $Y^{ROI}\in\{0, 1\}^{R \times C}$, $Y^{ROI}_{i, j}$ indicates the ($i$, $j$) -th element of Y, where $Y^{ROI}_{i ,j}$ = 0 denotes AU $j$ is inactive in AU ROI $i$, and $Y^{ROI}_{i, j}$ = 1 denotes AU $j$ is active in AU ROI $i$. In addition, the ground truth $Y^{ROI}$ must satisfy the constraint of AU region partition rule: $Y^{ROI}_{i, j}$ = 0 if AU $j$ does not belong to the $i$-th AU ROI. In particular, when an AU ROI consists of multiple AUs, just like the ROI containing AU12, AU14 and AU15 in BP4D, the $Y^{ROI}_{i, j}$ also follows the above rules. The ROI-level label also helps to improve AU detection performance through the space constraint and supervised information of the ROIs.  
Finally, the overall loss function for AU deep representation extraction is shown below:
\begin{equation}
    L_{ROI} = L_{ROI\_softmax} + \lambda _{1}L_{R}
\end{equation}
\noindent
in which, $\lambda_{1}$ is a trade-off parameter.

\vspace{-3mm}
\subsection{AU Relation Graph}
\vspace{-3mm}
In this section, AU relation graph is proposed to encoder those AU relations, in which, AUs with high confidence of relations are connected together. The relationships among AUs in the ROIs are analyzed to construct the AU relation graph. In the graph, we connect pairs of related AU ROIs together. The graph will show in Section 4.

Formally, we assume the number of the AU is $C$, the number of AU ROI is $R$. Given all labels in the training-set, the conditional probability that AU $j$ equals 1 when AU $i$ equals 1 is calculated. Relation matrix $M$ of C$\times$C dimension is obtained, and then in order to transform $M$ into symmetric matrix $M^{sym}$, the following function is introduced as:

\vspace{-2mm}
\begin{equation}
M^{sym}_{ij} = M_{ij} + M_{ji}
\end{equation}
\vspace{-2mm}

where $M^{ij}$ denotes the (i,j)-th element of the matrix $M$. Then, a threshold $h$ is set to convert $M^{sym}_{ij}$ into a 0-1 matrix $M^{bool}$ as following:

\vspace{-2mm}
\begin{equation}
    M^{bool}_{ij} = \left\{\begin{matrix}
    1 &if\ M^{sym}_{ij} >= threshold\\ 
    0 &if\ M^{sym}_{ij} < threshold
\end{matrix}\right.
\end{equation}
\vspace{-3mm}
\\
\indent
In the next, graph $G$ with $R$ AU ROI nodes is built according $M^{bool}$, in which, $R$ is the number of AU ROIs. Firstly, the node in $G$ is connected to itself. Secondly, each node is connected with its symmetrical node, i.e.: AU1\_left ROI and AU1\_right ROI,  AU23\_up and AU23\_down. Thirdly, if $M^{bool}_{ij}$ = 1, it shows that AU $i$ is strongly related to AU $j$, so these nodes belonging to AU $i$ are connected to those nodes belonging to AU $j$. Finally, the last node representing the whole facial image is connected to all nodes, which lets the global feature help the local features learn more AU information. By building the AU relation graph $G$, we can obtain richer AU relation information and enlarge the ability of classifiers in subsequent inference process.

\vspace{-3mm}
\subsection{Convolutions on Graph $G$}
\vspace{-2mm}
To perform reasoning on the graph, we apply the Graph Convolutional Networks (GCNs) proposed in~\cite{kipf2016semi}. Different from standard convolutions which operates on a local regular grid, the graph convolutions allow us to compute the response of a node based on its neighbors defined by the graph relations. Thus performing graph convolutions is equal to performing message passing inside the graphs. The outputs of the GCNs are updated features of each ROI node. 
Inspired by the above, we design a GCN-based multi-label encoder for AU detection. We can represent one layer of graph convolution as:
\vspace{-2mm}
\begin{equation}
    Z_{i+1}=G \times Z_{i} \times W_{i}
\end{equation}
\vspace{-3mm}
\\
\indent
where $G$ represents the adjacency graph we have introduced above with R$\times$R dimension. $Z_i$ denotes the input features in the $i$-th graph convolution operation, in particular, $Z_0$ denotes the latent vectors with $R \times d_0$, and $W_i$ is the weight matrix of the layer. $W_0$ with dimension $d_0 \times d_1$, $W_1$ with dimension $d_1 \times d_2$. Thus, the output of two graph convolution layers $Z_2$ is in $R \times d_2$ dimension. After each layer of graph convolutions, we apply two functions including the Dropout and then ReLU before the updated feature $Z_2$ is forwarded to the next layer. \\

\vspace{-3mm}
\subsection{Facial AU Detection:}
\vspace{-2mm}
As illustrated in Figure~\ref{fig1}, the updated feature $Z_2$ after graph convolutions is flatten. Then, the flatten feature is forwarded to a fully connected network (FCN) for AU detection. Finally, we get the detection results $\Hat{Y}$ with $C$-dim.

Facial AU detection can be regarded as a multi-label binary classification problem with the following weighted multi-label softmax loss $L_{softmax}$:
\vspace{-3mm}
\begin{equation}
    L_{softmax}(Y, \hat{Y}) = -\frac{1}{C}\sum_{i=1}^{C}w_{i}\left [ Y_{i}log\hat{Y}_{i} + (1 - Y_{i})log(1 - \hat{Y}_{i}) \right ]
\end{equation}
\vspace{-3mm}
\\
\indent
where $Y_i$ denotes the ground-truth probability of occurrence for the $i$-th AU,
which is $1$ if occurrence and $0$ otherwise, and $\Hat{Y}_i$ denotes the corresponding predicted occurrence probability for the $i$-th AU. The trade-off weight $w_i$ is introduced to alleviate the data imbalance problem. For most facial AU detection benchmarks, the occurrence rates of AUs are imbalanced~\cite{liu2018conditional,martinez2017automatic}. Since AUs are not mutually independent, imbalanced training data has a bad influence on this multi-label learning task. Particularly, we set $w_{i}$ = $\frac{(1/r_{i})C}{\sum_{i=1}^{C}(1/r_{i})}$ ,where $r_i$ is the occurrence rate of the $i$-th AU in the training set.
\\
\indent
In some cases, some AUs appear rarely in training samples, for which the softmax loss often makes the network prediction strongly biased towards absence. To overcome this limitation, a weighted multi-label Dice coefficient loss $L_{dice}$~\cite{milletari2016v} is further introduced as following:
\vspace{-3mm}
\begin{equation}
    L_{dice}(Y, \hat{Y}) = \frac{1}{C}\sum_{i=1}^{C}w_{i}(1-\frac{2Y_{i}\hat{Y}_{i}+\epsilon }{Y_{i}^{2}+\hat{Y}_{i}^{2}+\epsilon })
\end{equation}
\vspace{-3mm}
\\
\indent
where $\epsilon$ is the smooth term. Dice coefficient is also known as F1-score: $F1 = 2pr/(p + r)$, the most popular metric for facial AU detection, where $p$ and $r$ denote precision and recall respectively. With the help of the weighted Dice coefficient loss, we also take into account the consistency between the learning process and the evaluation metric. Finally, the AU detection loss is defined as:
\vspace{-3mm}
\begin{equation}
    L_{au} = L_{softmax} + \lambda _{2}L_{dice}
\end{equation}
\vspace{-3mm}
where $\lambda _{2}$ is a trade-off parameter.

\vspace{-3mm}
\section{Experiments}
\vspace{-3mm}
\subsection{Setting}\label{SCM}
\indent
$\textbf{Dataset:}$ The effectiveness of our proposed AU-GCN is evaluated on two benchmark datasets: BP4D~\cite{zhang2013high} and DISFA~\cite{mavadati2013disfa}. For BP4D and DISFA, a 3-fold partition is adopted to ensure subjects were mutually exclusive in train/val/test sets by following previous related work~\cite{zhao2016deep,li2017eac}. The frames with intensities equal or greater than 2 are considered as positive, while others are treated as negative. BP4D contains 2D and 3D videos of 41 young adults during various emotion inductions while interacting with an experimenter. We used 328 videos (41 participants$\times$8 videos each) with 10 AUs coded, resulting in $\sim$140,000 valid face images. For each AU, we sampled 100 positive frames and 200 negative frames for each video. DISFA~\cite{mavadati2013disfa} contains 27 subjects watching video clips, and provides 8 AU annotations with intensities. There were $\sim$130,000 valid face images. We used the frames with AU intensities with 2 or higher as positive samples, and the rest as negative ones. To be consistent with the 8-video setting of BP4D, we sampled 800 positive frames and 1600 negative frames for each video.\\
\indent
$\textbf{Metrics:}$ The AU detection performance was evaluated on two commonly used frame-based metrics: F1-score and area under curve (AUC). F1-score is the harmonic mean of precision and recall, and widely used in AU detection. AUC quantifies the relation between true and false positives. For each method, we computed average metrics over all AUs (denoted as Avg.).\\
\indent
$\textbf{Implementation:}$ For each face image, we perform similarity transformation to obtain a 200$\times$200$\times$3 color face. This transformation is shape-preserving and brings no change to the expression. In order to enhance the diversity of training data, the face images are flipped for data augmentation. Our AU-GCN is trained using PyTorch with stochastic gradient descent(SGD), a mini-batch size of 256, a momentum of 0.9 and a weight decay of 0.0005. We decay the learning rate by 0.1 after every 10 epochs. The structure parameters of AU-GCN are chosen as $d_0$ = 150, $d_1$ = 30, $d_2$ = 12 , $n$ is 25, $C$ is $12$ for BP4D and $8$ for DISFA, $R$ is 19 for BP4D and $14$ for DISFA. The graph connection matrix $M^{bool}$ on BP4D and DISFA are shown in Table~\ref{Tab_G_Bp4D_DISFA}. The hyperparameters  $\lambda_1$, $\lambda_2$ are obtained by cross validation. In our experiments, set  $\lambda_1$ = 3 and $\lambda_2$ = 4. AU-GCN is firstly trained with AE optimized with 12 epochs. Next, we read the parameters before getting the $R$ latent vectors and train with all the modules optimized with 40 epochs. 

\vspace{-3mm}
\begin{table}[htb]
\caption{Graph Connection Matrix $M^{bool}$ on BP4D and DISFA}
\centering
\scriptsize
\subtable[BP4D]{
       \begin{tabular}{|c|c|c|c|c|c|c|c|c|c|c|c|c|}
\cline{1-13} 
\textbf{AU} & \textbf{1}& \textbf{2}& \textbf{4}& \textbf{6}&\textbf{7}&\textbf{10}&\textbf{12}&\textbf{14}&\textbf{15}&\textbf{17}&\textbf{23}&\textbf{24}\\
\hline\hline
\textbf{1}&1&1& 0& 0& 0& 1& 0& 0& 0& 0& 0&0 \\
\hline
\textbf{2}& 1& 1& 0& 0& 0& 1& 1& 0& 0& 0& 0& 0\\
\hline
\textbf{4}& 0& 0& 1& 0& 1& 0& 0& 0& 0& 0& 0& 0\\
\hline
\textbf{6}& 0& 0& 0& 1& 1& 1& 1& 1& 1& 1& 0& 0\\
\hline
\textbf{7}& 0& 0& 1& 1& 1& 1& 1& 1& 1& 1& 1& 0\\
\hline
\textbf{10}& 1& 1& 0& 1& 1& 1& 1& 1& 1& 1& 1& 0\\
\hline
\textbf{12}& 0& 1& 0& 1& 1& 1& 1& 1& 1& 1& 1& 0\\
\hline
\textbf{14}& 0& 0& 0& 1& 1& 1& 1& 1& 1& 1& 1& 1\\
\hline
\textbf{15}& 0& 0& 0& 1& 1& 1& 1& 1& 1& 1& 0& 0\\
\hline
\textbf{17}& 0& 0& 0& 1& 1&1 &1 & 1& 1& 1&1 &1 \\
\hline
\textbf{23}& 0& 0& 0& 0& 1& 1& 1& 1& 0& 1&1 &0 \\
\hline
\textbf{24}& 0& 0& 0& 0& 0& 0& 0& 1& 0& 1& 0& 1\\
\cline{1-13} 
 
\end{tabular}
       \label{Tab_G_BP4D}
}
\qquad
\subtable[DISFA]{        
       \begin{tabular}{|c|c|c|c|c|c|c|c|c|}
\cline{1-9} 
\textbf{AU} & \textbf{1}& \textbf{2}& \textbf{4}& \textbf{6}&\textbf{9}&\textbf{12}&\textbf{25}&\textbf{26}\\
\hline\hline
\textbf{1}& 1& 1& 0& 0& 0& 0& 0& 0\\
\hline
\textbf{2}& 1& 1& 0& 0& 0& 0& 0& 0\\
\hline
\textbf{4}& 0& 0& 1& 0& 1& 0& 0& 0\\
\hline
\textbf{6}& 0& 0& 0& 1& 0& 1& 1& 0\\
\hline
\textbf{9}& 0& 0& 1& 0& 1& 0& 0& 0\\
\hline
\textbf{12}& 0& 0& 0& 1& 0& 1& 1& 0\\
\hline
\textbf{25}& 0& 0&0 & 1& 0& 1& 1& 1\\
\hline
\textbf{26}& 0& 0& 0& 0&0 & 0& 1& 1\\

\cline{1-9} 
 
\end{tabular}
       \label{Tab_G_DISFA}
}
\label{Tab_G_Bp4D_DISFA}
\end{table}

\vspace{-2mm}
\subsection{Comparison with State-of-the-Art Methods}
\vspace{-2mm}

We compare our method AU-GCN against state-of-the-art single-image based AU detection works under the same 3-fold cross validation setting. These methods include both traditional methods, LSVM~\cite{fan2008liblinear}, JPML~\cite{zhao2016joint}, and deep learning methods, LCN~\cite{taigman2014deepface}, DRML~\cite{zhao2016deep} and EAC-Net~\cite{li2017eac}. Note that EAC-Net~\cite{li2017eac} is not compared AUC due to its metrics of accuracy instead of AUC.

\begin{table*}[!htbp]
\caption{F1-score and AUC for 12 AUs on BP4D}
\scriptsize
\begin{center}
\begin{tabular}{|c|c|c|c|c|c|c|c|c|c|c|c|}
\hline
\textbf{}&\multicolumn{6}{|c|}{\textbf{F1-score}}&\multicolumn{5}{|c|}{\textbf{AUC}} \\
\cline{1-12} 
\textbf{AU} & \textbf{LSVM}& \textbf{JPML}& \textbf{LCN}& \textbf{DRML}&\textbf{EAC-Net}&\textbf{AU-GCN}& \textbf{LSVM}& \textbf{JPML}& \textbf{LCN}& \textbf{DRML}&\textbf{AU-GCN}\\
\hline\hline
1& 23.2&32.6  &45.0 &36.4 &39.0 &\textbf{46.8}&20.7  &40.7  &51.9  &55.7  & \textbf{58.3}   \\
\hline
2&22.8 & 25.6 &41.2  &\textbf{41.8} & 35.2 & 38.5 &17.7  &42.1  &50.9  &54.5  & \textbf{71.2}   \\
\hline
4 &23.1  &37.4  &42.3  &43.0 &48.6  &\textbf{60.1} &22.9  &46.2  &53.6  &58.8  & \textbf{85.6}  \\
\hline
6 & 27.2 & 42.3 & 58.6 & 55.0 &76.1 &\textbf{80.1} &20.3  &40.0  &53.2  &56.6  & \textbf{89.7}  \\
\hline
7 & 47.1 & 50.5 & 52.8 & 67.0 &72.9 &\textbf{79.5} &44.8  &50.0  &63.7  &61.0  & \textbf{83.5}  \\
\hline
10 &77.2  & 72.2 & 54.0 & 66.3 &81.9 &\textbf{84.8} &73.4  &75.2  &62.4  &53.6  & \textbf{86.7}  \\
\hline
12 &63.7  &74.1  &54.7  &65.8 &86.2 &\textbf{88.0} &55.3  &60.5  &61.6  &60.8  & \textbf{93.5}  \\
\hline
14 & 64.3 & 65.7 & 59.9 & 54.1 &58.8 &\textbf{67.3} &46.8  &53.6  &58.8  &57.0  & \textbf{78.0}  \\
\hline
15 &18.4  &38.1  & 36.1 & 33.2 &37.5 &\textbf{52.0} &18.3  &50.1  &49.9  &56.2  & \textbf{86.6}  \\
\hline
17 & 33.0 & 40.0 & 46.6 & 48.0 &59.1 &\textbf{63.2} &36.4  &42.5  &48.4  &50.0  & \textbf{81.2}  \\
\hline
23 & 19.4 & 30.4 & 33.2 & 31.7 &35.9 &\textbf{40.9} &19.2  &51.9  &50.3  &53.9  & \textbf{80.3}  \\
\hline
24 & 20.7 & 42.3 & 35.3 & 30.0 &35.8 &\textbf{52.8} &11.7  &53.2  &47.7  &53.9  & \textbf{91.4}  \\
\hline
Avg & 35.3 & 45.9 & 46.6 & 48.3 &55.9 &\textbf{62.8}&32.2  &50.5  &54.4  &56.0  & \textbf{87.3}\\ 
\cline{1-12} 
 
\end{tabular}
\label{tab1}
\end{center}
\end{table*}

Table \ref{tab1} reports the F1-score and AUC results of different methods on BP4D. It can be seen that our AU-GCN outperforms all these previous works on the challenging BP4D dataset. AU-GCN is superior to all the conventional methods, which demonstrates the strength of deep learning based methods. Compared to the state-of-the-art methods, AU-GCN brings significant relative increments of 6.9\% and 31.3\% respectively for average F1-score and AUC, which verifies the effectiveness of AU relation modeling with GCN. In addition, our method obtains high accuracy without sacrificing F1-score, which is attributed to the integration of the softmax loss and the Dice coefficient loss.

\begin{table*}[!htbp]
\scriptsize 
\caption{F1-score and AUC for 8 AUs on DISFA}
\begin{center}
\begin{tabular}{|c|c|c|c|c|c|c|c|c|c|c|c|}
\hline
\textbf{}&\multicolumn{6}{|c|}{\textbf{F1-score}}&\multicolumn{5}{|c|}{\textbf{AUC}} \\

\cline{1-12} 
\textbf{AU} & \textbf{LSVM}& \textbf{APL}& \textbf{LCN}& \textbf{DRML}&\textbf{EAC-Net}&\textbf{AU-GCN} & \textbf{LSVM}& \textbf{APL}& \textbf{LCN}& \textbf{DRML}&\textbf{AU-GCN}\\

\hline\hline
1 &10.8  &11.4  & 12.8 &17.3 &\textbf{41.5}  &32.3 & 21.6 & 32.7 & 44.1 & \textbf{53.3} &47.1  \\
\hline
2 & 10.0 &12.0  & 12.0 & 17.7 &\textbf{26.4} &19.5 & 15.8 & 27.8 & 52.4 & 53.2 &\textbf{61.1}  \\
 \hline
4 & 21.8 & 30.1 & 29.7 & 37.4 &\textbf{66.4} &55.7 & 17.2 & 37.9 & 47.7 &60.0  &\textbf{72.5}\\
 \hline
6 & 15.7 & 12.4 & 23.1 & 29.0 &50.7 &\textbf{57.9} & 8.7 &13.6  &39.7  & 54.9 &\textbf{77.9} \\
 \hline
9 & 11.5 & 10.1 & 12.4 & 10.7 &\textbf{80.5} &61.4 &15.0  &\textbf{64.4}  & 40.2 &51.5  &62.3\\
 \hline
12 & 70.4 & 65.9 & 26.4 & 37.7 &\textbf{89.3} &62.7  & 93.8 & \textbf{94.2} &54.7  & 54.6 &91.6 \\
 \hline
25 & 12.0 & 21.4 & 46.2 & 38.5 &88.9 &\textbf{90.9} & 3.4 &50.4  & 48.6 & 48.6 &\textbf{95.8}\\
 \hline
26 & 22.1 & 26.9 & 30.0 & 20.1 &15.6 &\textbf{60.0} & 20.1 &47.1  & 47.0 & 45.3 &\textbf{88.4}\\
 \hline
Avg & 21.8 & 23.8 &24.0  & 26.7 &48.5 & \textbf{55.0}&27.5  & 46.0 & 46.8 &52.3  & \textbf{74.6}\\

\cline{1-12}
\end{tabular}
\label{tab3}
\end{center}
\end{table*}





Experimental results on DISFA dataset are shown in Table~\ref{tab3}, from which it can be observed that our AU-GCN outperforms all the state-of-the-art works with even more significant improvements. Specifically, AU-GCN increases the average F1-score and AUC relatively by 6.5\% and 22.3\% over the state-of-the-art methods, respectively. Due to the serious data imbalance issue in DISFA, performances of different AUs fluctuate severely in most of the previous methods. For instance, the accuracy of AU 12 is far higher than that of other AUs for LSVM and APL. Although both AU-GCN and EAC-Net use the AU local features, the GCN better expresses the AU relation information.

\vspace{-3mm}
\subsection{Ablation Study}
\vspace{-3mm}
To investigate the effectiveness of each component in our framework, Table~\ref{tab5} present the average F1-score and AUC for different variants of AU-GCN on BP4D benchmark, where ``w/o'' is the abbreviation of ``without''. Each variant is composed by different components of our framework. AU-Net is the framework without GCN relation modeling (GCN), Dice loss (D) and global facial information (F).

\vspace{-4mm}
\begin{table}[htbp]
\caption{F1-score on BP4D for Ablation Study}
\scriptsize
\begin{center}
\begin{tabular}{|c|c|c|c|c|c|c}
\hline

\textbf{Methods} & \textbf{GCN}& \textbf{D}& \textbf{F}& \textbf{F1-score}&\textbf{AUC}\\

\hline\hline
AU-Net &$\times$  &$\times$  &$\times$  & 54.0 &81.3  \\
\hline
AU-Net+D &$\times$  &$\checkmark$ &$\times$  &56.4  &83.7  \\
 \hline
AU-GCN w/o F,D &$\checkmark$  &$\times$  &$\times$  &59.1 &83.9   \\
 \hline
AU-GCN w/o F &$\checkmark$  &$\checkmark$  & $\times$ &61.5  &85.8  \\
 \hline
AU-GCN w/o D &$\checkmark$  &$\times$  &$\checkmark$  &61.9 &86.8   \\
 \hline
AU-GCN &$\checkmark$  &$\checkmark$  &$\checkmark$  & \textbf{62.8} &\textbf{87.3} \\

\cline{1-6}
\end{tabular}
\label{tab5}
\end{center}
\end{table}
\vspace{-3mm}


\vspace{-3mm}
$\textbf{Contribution of the GCN:}$ By integrating the graph convolutional network (GCN), AU-GCN w/o F,D achieves higher F1-score and AUC results than AU-Net. In particular, the AU-Net is to concatenate $R$ node features and put the concatenated features into FCN instead of GCN. This result illuminates that GCN can capture the strong relationship between AUs, and strength the relation learning for AU detection.
\\
\indent
$\textbf{Integrating of whole facial information:}$ By adding the resized whole facial image as a node to GCN and connecting this node with all the other nodes, AU-GCN w/o D achieves better F1-score and AUC results compared to AU-GCN w/o F,D. Since the previous features are all local AU features, benefiting from the whole facial image to add global feature for GCN, our method obtains more significant performance, which demonstrates that global facial information is helpful for local AU detection.
\\
\indent
$\textbf{Integrating of Dice loss:}$ After integrating the weighted softmax loss with the Dice loss, AU-GCN w/o F attains higher average F1-score and AUC than AU-GCN w/o F,D. The softmax loss focus more on the classification accuracy, rather than the balance between precision and recall, the Dice loss which optimizes the network from the perspective of F1-score, so this loss can make F1-score and AUC achieve good results.

\vspace{-3mm}
\section{Conclusion}
\vspace{-3mm}
In this paper, to makes full use of AU local features and their relationship, we have presented a facial AU detection methods by integrating graph convolution network for explicit AU relation modeling. To the best of our knowledge, this is the first study that combines facial AU detection and GCN with one end-to-end framework. Extensive experiments on two benchmark AU datasets demonstrate that the proposed network outperformed state-of-the-art methods for AU detection, the effectiveness of the proposed modules in the framework are also validated through a series of ablation study.

\vspace{-3mm}
\section*{Acknowledgements}
\vspace{-3mm}
This work is supported by the National Natural Science Foundation of China under Grants of 41806116 and 61503277. We gratefully acknowledge the support of NVIDIA Corporation with the donation of the Titan V GPU used for this research.

\vspace{-3mm}
\bibliographystyle{splncs04}
\bibliography{refer}

\begin{thebibliography}{10}
\providecommand{\url}[1]{\texttt{#1}}
\providecommand{\urlprefix}{URL }
\providecommand{\doi}[1]{https://doi.org/#1}

\bibitem{atwood2016diffusion}
Atwood, J., Towsley, D.: Diffusion-convolutional neural networks. In: Advances
  in Neural Information Processing Systems. pp. 1993--2001 (2016)

\bibitem{bruna2013spectral}
Bruna, J., Zaremba, W., Szlam, A., LeCun, Y.: Spectral networks and locally
  connected networks on graphs. arXiv preprint arXiv:1312.6203  (2013)

\bibitem{defferrard2016convolutional}
Defferrard, M., Bresson, X., Vandergheynst, P.: Convolutional neural networks
  on graphs with fast localized spectral filtering. In: Advances in neural
  information processing systems. pp. 3844--3852 (2016)

\bibitem{duvenaud2015convolutional}
Duvenaud, D.K., Maclaurin, D., Iparraguirre, J., Bombarell, R., Hirzel, T.,
  Aspuru-Guzik, A., Adams, R.P.: Convolutional networks on graphs for learning
  molecular fingerprints. In: Advances in neural information processing
  systems. pp. 2224--2232 (2015)

\bibitem{ekman1997face}
Ekman, R.: What the face reveals: Basic and applied studies of spontaneous
  expression using the Facial Action Coding System (FACS). Oxford University
  Press, USA (1997)

\bibitem{fan2008liblinear}
Fan, R.E., Chang, K.W., Hsieh, C.J., Wang, X.R., Lin, C.J.: Liblinear: A
  library for large linear classification. Journal of machine learning research
   \textbf{9}(Aug),  1871--1874 (2008)

\bibitem{hammond2011wavelets}
Hammond, D.K., Vandergheynst, P., Gribonval, R.: Wavelets on graphs via
  spectral graph theory. Applied and Computational Harmonic Analysis
  \textbf{30}(2),  129--150 (2011)

\bibitem{henaff2015deep}
Henaff, M., Bruna, J., LeCun, Y.: Deep convolutional networks on
  graph-structured data. arXiv preprint arXiv:1506.05163  (2015)

\bibitem{kipf2016semi}
Kipf, T.N., Welling, M.: Semi-supervised classification with graph
  convolutional networks. arXiv preprint arXiv:1609.02907  (2016)

\bibitem{li2019semantic}
Li, G., Zhu, X., Zeng, Y., Wang, Q., Lin, L.: Semantic relationships guided
  representation learning for facial action unit recognition. arXiv preprint
  arXiv:1904.09939  (2019)

\bibitem{li2017eac}
Li, W., Abtahi, F., Zhu, Z., Yin, L.: Eac-net: A region-based deep enhancing
  and cropping approach for facial action unit detection. In: 2017 12th IEEE
  International Conference on Automatic Face \& Gesture Recognition (FG 2017).
  pp. 103--110. IEEE (2017)

\bibitem{li2015gated}
Li, Y., Tarlow, D., Brockschmidt, M., Zemel, R.: Gated graph sequence neural
  networks. arXiv preprint arXiv:1511.05493  (2015)

\bibitem{liu2018conditional}
Liu, Z., Song, G., Cai, J., Cham, T.J., Zhang, J.: Conditional adversarial
  synthesis of 3d facial action units. arXiv preprint arXiv:1802.07421  (2018)

\bibitem{martinez2017automatic}
Martinez, B., Valstar, M.F., Jiang, B., Pantic, M.: Automatic analysis of
  facial actions: A survey. IEEE transactions on affective computing  (2017)

\bibitem{masci2011stacked}
Masci, J., Meier, U., Cire{\c{s}}an, D., Schmidhuber, J.: Stacked convolutional
  auto-encoders for hierarchical feature extraction. In: International
  Conference on Artificial Neural Networks. pp. 52--59. Springer (2011)

\bibitem{mavadati2013disfa}
Mavadati, S.M., Mahoor, M.H., Bartlett, K., Trinh, P., Cohn, J.F.: Disfa: A
  spontaneous facial action intensity database. IEEE Transactions on Affective
  Computing  \textbf{4}(2),  151--160 (2013)

\bibitem{milletari2016v}
Milletari, F., Navab, N., Ahmadi, S.A.: V-net: Fully convolutional neural
  networks for volumetric medical image segmentation. In: 2016 Fourth
  International Conference on 3D Vision (3DV). pp. 565--571. IEEE (2016)

\bibitem{ng2018bayesian}
Ng, Y.C., Colombo, N., Silva, R.: Bayesian semi-supervised learning with graph
  gaussian processes. In: Advances in Neural Information Processing Systems.
  pp. 1683--1694 (2018)

\bibitem{niepert2016learning}
Niepert, M., Ahmed, M., Kutzkov, K.: Learning convolutional neural networks for
  graphs. In: International conference on machine learning. pp. 2014--2023
  (2016)

\bibitem{song2015exploiting}
Song, Y., McDuff, D., Vasisht, D., Kapoor, A.: Exploiting sparsity and
  co-occurrence structure for action unit recognition. In: 2015 11th IEEE
  international conference and workshops on automatic face and gesture
  recognition (FG). vol.~1, pp.~1--8. IEEE (2015)

\bibitem{taigman2014deepface}
Taigman, Y., Yang, M., Ranzato, M., Wolf, L.: Deepface: Closing the gap to
  human-level performance in face verification. In: Proceedings of the IEEE
  conference on computer vision and pattern recognition. pp. 1701--1708 (2014)

\bibitem{8510873}
{Wang}, S., {Hao}, L., {Ji}, Q.: Facial action unit recognition and intensity
  estimation enhanced through label dependencies. IEEE Transactions on Image
  Processing  \textbf{28}(3),  1428--1442 (March 2019).
  \doi{10.1109/TIP.2018.2878339}

\bibitem{Wang2014Capturing}
Wang, Z., Li, Y., Wang, S., Qiang, J.: Capturing global semantic relationships
  for facial action unit recognition. In: IEEE International Conference on
  Computer Vision (2014)

\bibitem{zhang2013high}
Zhang, X., Yin, L., Cohn, J.F., Canavan, S., Reale, M., Horowitz, A., Liu, P.:
  A high-resolution spontaneous 3d dynamic facial expression database. In: 2013
  10th IEEE International Conference and Workshops on Automatic Face and
  Gesture Recognition (FG). pp.~1--6. IEEE (2013)

\bibitem{zhao2016joint}
Zhao, K., Chu, W.S., De~la Torre, F., Cohn, J.F., Zhang, H.: Joint patch and
  multi-label learning for facial action unit and holistic expression
  recognition. IEEE Transactions on Image Processing  \textbf{25}(8),
  3931--3946 (2016)

\bibitem{zhao2016deep}
Zhao, K., Chu, W.S., Zhang, H.: Deep region and multi-label learning for facial
  action unit detection. In: Proceedings of the IEEE Conference on Computer
  Vision and Pattern Recognition. pp. 3391--3399 (2016)

\bibitem{Zou2005A}
Zou, M., Conzen, S.D.: A new dynamic bayesian network (dbn) approach for
  identifying gene regulatory networks from time course microarray data.
  Bioinformatics  \textbf{21}(1),  71--79 (2005)

\end{thebibliography}
%




\end{document}